\documentclass[lettersize,journal]{IEEEtran}
\usepackage{amsmath,amsfonts}
\usepackage{algorithmic}
\usepackage{algorithm}
\usepackage{array}
\usepackage[caption=false,font=normalsize,labelfont=sf,textfont=sf]{subfig}
\usepackage{textcomp}
\usepackage{stfloats}
\usepackage{url}
\usepackage{verbatim}
\usepackage{graphicx}
\usepackage{cite}

\usepackage{xcolor}
\usepackage{hyperref}
\hypersetup{
    colorlinks=true,
    linkcolor=purple,
    filecolor=magenta,      
    urlcolor=purple,
}
\usepackage{url}            
\usepackage{booktabs}       
\usepackage{xcolor}         
\usepackage{amssymb}
\usepackage{makecell}
\newcommand{\etal}{\textit{et al}.}
\newcommand{\ie}{\textit{i}.\textit{e}.}
\newcommand{\eg}{\textit{e}.\textit{g}.}

\usepackage{pifont}

\usepackage{arydshln}
\usepackage{wrapfig}
\usepackage{multirow}
\usepackage{tabularx}

\begin{document}

\title{TMT: Tri-Modal Translation between\\Speech, Image, and Text by Processing\\Different Modalities as Different Languages}

\author{Minsu Kim$^{*}$, Jee-weon Jung$^{*}$, Hyeongseop Rha, Soumi Maiti,\\ Siddhant Arora, Xuankai Chang, Shinji Watanabe, Yong Man Ro
\thanks{$^*$Equal Contribution.}
\thanks{This work was supported by the National Research Foundation of Korea (NRF) grant funded by the Korea government (MSIT) (No.~NRF-2022R1A2C2005529).}
\thanks{M. Kim, H. Rha, and Y. M. Ro are with the Integrated Vision and Language Lab., School of Electrical Engineering, Korea Advanced Institute of Science and Technology (KAIST), 291 Daehak-ro, Yuseong-gu, Daejeon, 34141, Republic of Korea (e-mail: \{ms.k; ryool\_1832; ymro\}@kaist.ac.kr).} 
\thanks{J. Jung, S. Maiti, S. Arora, X. Chang, and S. Watanabe are with the Language Technologies Institute, Carnegie Mellon University, 5000 Forbes Ave, Pittsburgh, PA 15213, USA (e-mail: \{jeeweonj; smaiti; siddhana; xuankaic; swatanab\}@andrew.cmu.edu). 
Corresponding author: Y. M. Ro (fax: 82-42-350-5494).}}



\maketitle

\begin{abstract}
The capability to jointly process multi-modal information is becoming essential. However, the development of multi-modal learning is hindered by the substantial computational requirements and the limited availability of paired multi-modal data. We propose a novel Tri-Modal Translation (TMT) model that translates between arbitrary modalities spanning speech, image, and text. We introduce a simple yet efficient and effective approach, treating speech and image modalities as discrete text modality and approaching multi-modal translation as a well-established machine translation problem. To this end, we tokenize speech and image data into discrete tokens, resulting in a significant reduction in computational cost. Furthermore, by incorporating back translation into multi-modal translation, unpaired data can also be utilized for training. TMT can perform six modality translation tasks and consistently outperforms its single-model counterparts. TMT significantly reduces the required data size (in bits) for training, to approximately 0.2\% for speech data and 0.04\% for image data, respectively.
\end{abstract}


\section{Introduction}
\label{sec:intro}
\IEEEPARstart{I}{n} search of artificial general intelligence, seamlessly processing and representing multi-modal information is a crucial prerequisite. Leveraging numerous efforts on the investigation of diverse multi-modal work \cite{gu2020toward,chen2023building}, systems capable of processing audio-visual~\cite{chen2021localizing,liu2022dense,hong2023watch}, audio-text~\cite{yang2023diffsound,wu2023improving}, and text-visual~\cite{ramesh2021dalle,li2023gligen} modalities are actively emerging. Furthermore, developing a unified model that processes tri-modalities (\ie, audio, visual, and text) is drawing attention given its potential for advancing general intelligence~\cite{wu2023next,han2023imagebind}.

However, building such a unified model remains challenging due to multifaceted factors. Essentially, the data size (in bits) of speech and image modalities exceeds that of text by a significant margin. Furthermore, the distinct nature of different modalities intrinsically burdens the multi-modal systems to include two roles: processing modality-specifics and translating between different modalities. Consequently, the computational load required to train multi-modal models usually exceeds that of uni-modal models, attributed to the large data size and the effort necessary to do two roles simultaneously\footnote{We believe this computational burden is one of the most impactful bottlenecks that hinder large multi-modal models, compared to Large Language Models (LLMs), where one can scale the model size and efficiently train it with text-only data.}. Furthermore, the necessity of paired multi-modal data is another obstacle because the amount of such data (\eg, speech-image-text) is less abundant than uni-modal data. 

In this paper, we propose a novel Tri-Modal Translation (TMT) framework that freely translates between different modalities (\ie, speech, image, and text) using a single model. Specifically, we present a novel viewpoint on multi-modal modeling by interpreting different modalities as text modality with different languages. We split the Multi-Modal Translation (MMT) problem into modality-specific processings and an integrated translation, where the latter is treated as the Neural Machine Translation (NMT). To this end, we discretize all modalities into tokens in advance\footnote{Image and speech are represented as sequences of integers.} by employing pre-trained modality-specific tokenizers. This enables us to leverage the expertise from each field, discarding modality-specifics from the core translation modeling.

The proposed TMT framework and viewpoint introduce several benefits. First, the proposed system is ``purely multi-modal'' by uniformly interfacing all modalities together different to the method of mapping different modalities into the text space of LLM~\cite{wu2023next}, which we believe is the future direction of building multi-modal LLMs. Second, through discretization, the computational burden in modeling speech and image is significantly relaxed, specifically by more than $99$\%~\cite{chang2023exploration,kim2024towards} in the aspect of required bits. Finally, we can employ unpaired training data through Back Translation (BT)~\cite{sennrich2016backtranslation}, a well-known data augmentation strategy in NMT\footnote{
BT typically translates target monolingual data into a source language to construct source-to-target parallel data.}. Therefore, even utilizing extra uni-modal data~\cite{deng2009imagenet,kahn2020librilight,fan2019eli5} in building TMT becomes possible. With the advantages of computational efficiency and the accessibility of unpaired data, multi-modal training holds the potential for scalable expansion.

The contributions of this paper can be summarized as follows:
1) To the best of our knowledge, this work is the first to explore translations between discretized (\ie, tokenized) tri-modalities of image, speech, and text. By handling multi-modal data with discretized tokens, we can greatly improve data efficiency, where the required bits for images are reduced to 0.035\% of their original size and 0.2\% for speech.
2) We treat the discretized modalities as text modality and regard them akin to distinct languages. By applying back translation, an NMT technique, we show that target uni-modal corpora can be also employed for MMT.
3) The proposed TMT encompasses direct speech-to-image synthesis and image-to-speech captioning, both of which have not been well addressed in the previous literature.
4) We show that the proposed TMT can be efficiently trained like text-only systems via extensive experiments and analysis on six MMT tasks. The pre-trained model, code, and qualitative results are available on \href{https://github.com/ms-dot-k/TMT}{github.com/ms-dot-k/TMT}.

\section{Related Work}
\label{sec:rel}
\subsection{Multi-Modal Translation (MMT)} 
Developing models capable of performing multiple MMT tasks is a popular research topic with diverse groups of researchers' investigation~\cite{kim2023many,ao2022speecht5,girdhar2023imagebind}. MUGEN~\cite{hayes2022mugen} explored the discretization of video, audio, and text obtained from synthetic game data, and successfully trained multiple MMT models capable of a single MMT task. VoxtLM~\cite{maiti2023voxtlm} proposed a decoder-only architecture to train a model that performs four MMT tasks ($2 \times 2$) spanning speech and text modalities. They discretized speech using a pre-trained self-supervised feature and treated discretized speech tokens in the same manner as text tokens within a combined vocabulary set. SEED~\cite{ge2023planting,ge2023making} proposed an image tokenizer using contrastive learning between image and text. With the learned image tokens, they finetuned an LLM spanning text and image modalities, covering two MMT tasks, image captioning and text-to-image synthesis. 
Different to VoxtLM and SEED, TMT is not restricted to bi-modal translation and can translate between tri-modalities. 

There also exist a few concurrent works aiming to handle tri-modalities.
NExT-GPT~\cite{wu2023next} tries to adapt different modalities into the text space of a pre-trained LLM by using several modality-specific encoders and adaptors, where the overall performance depends on the performance of the LLM. Distinct from NExT-GPT, our TMT is ``born multi-modal'' by learning from multi-modal data instead of projecting different modalities' representation into that of a text LLM. Gemini~\cite{team2023gemini} is a recently proposed multi-modal LLM. Different from Gemini, the proposed TMT can directly synthesize speech. Most significantly, different from the existing work, TMT interprets both discretized image and speech data as text tokens. Hence, the computational efficiency in handling multi-modal data is greatly improved. TMT successfully models all six MMT tasks with a unified interface with fully shared parameters.

\subsection{Image Captioning}
Image Captioning \cite{xu2015show} is a task to describe input image content in natural language. It has shown remarkable progress with the development of computer vision and natural language processing technologies. Recent methods showed that vision-language pre-training can significantly improve image captioning performances over training the model from scratch \cite{wang2022git}. For example, \cite{radford2021clip,jia2021align} proposed cross-modal contrastive learning, and \cite{li2021albef,kim2021vilt,li2022blip,li2023blip2} proposed multi-task pre-training which contains such as image-text matching, masked language modeling, and contrastive learning. Basically, the task can be regarded as an MMT problem of image-to-text translation.

Similar yet distinct from image captioning, image-to-speech captioning \cite{hsu2021text} aims to directly generate a spoken description of the input image. As the task does not go through with text modality, it is regarded as a challenging problem due to insufficient supervision and the complex nature of speech. Recent methods tried to mitigate the challenge by representing speech in pseudo-text format and showed promising results \cite{hsu2021text,effendi2021end,kim2024towards}. This task falls in text-to-speech translation.

\subsection{Language-driven Image Synthesis}
With the great development of visual codebook representation \cite{van2017neural,esser2021taming} and diffusion \cite{ho2020denoising,rombach2022ldm}, image generation has become more and more realistic. Especially, \cite{ramesh2021dalle,ramesh2022hierarchical,saharia2022photorealistic,chang2023muse,yang2024dmf} showed that generative models can now freely create images based on text descriptions by conditioning the text representations into the decoding process, such as diffusion process. Recently, \cite{koh2024generating} proposed a LLM-based text-to-image synthesis model that first trains with image captioning to enable the LLM to understand image representations. Different from them, the proposed TMT can perform 6 different tasks with a single trained model. Moreover, the proposed method handles the data efficiently through discretized image and speech tokens.

Different from text-based image synthesis, direct speech-to-image synthesis \cite{li2020direct,wang2020s2igan} has not well been addressed. Since many languages have no writing systems \cite{lee2022textless}, text-based systems cannot be applied to such languages. Hence, the demand for technology enabling spoken language-driven image synthesis, the speech-to-image translation, is expected to increase.

\subsection{Neural Machine Translation (NMT)} NMT aims to transform a source language sentence into a target language~\cite{stahlberg2020neural,song2021enhancing}. With the advent of advanced network architectures~\cite{kenton2019bert,lewis2020bart} and innovative training methodologies, such as denoising pre-training~\cite{ho2020denoising} and BT~\cite{sennrich2016backtranslation}, the field has rapidly evolved. Furthermore, NMT has expanded to encompass multi-modal systems, including speech-to-text machine translation~\cite{inaguma2020espnetst,li2021multilingual}, speech-to-speech machine translation~\cite{jia2019direct,zhang2021uwspeech}, and text-image machine translation~\cite{ma2023modal,ma2023multi}. Among them, BT aims to produce source-target language pairs by performing translation onto the monolingual target language data using an intermediate developing multilingual model. It follows the following steps in general: 1) train an intermediate model with supervised data. 2) produce the source-target language pairs by applying the intermediate model onto the target language monolingual data (\ie, applying BT). 3) by using both supervised data and back-translated data, further train the intermediate model to produce the final model. In this work, we treat discretized speech and images as pseudo-text so that TMT can be efficiently addressed using the format of NMT. In this viewpoint, NMT techniques, back translation and network architecture, are incorporated in TMT.

\begin{figure*}[t]
	\centering
	\centerline{\includegraphics[width=0.87\textwidth]{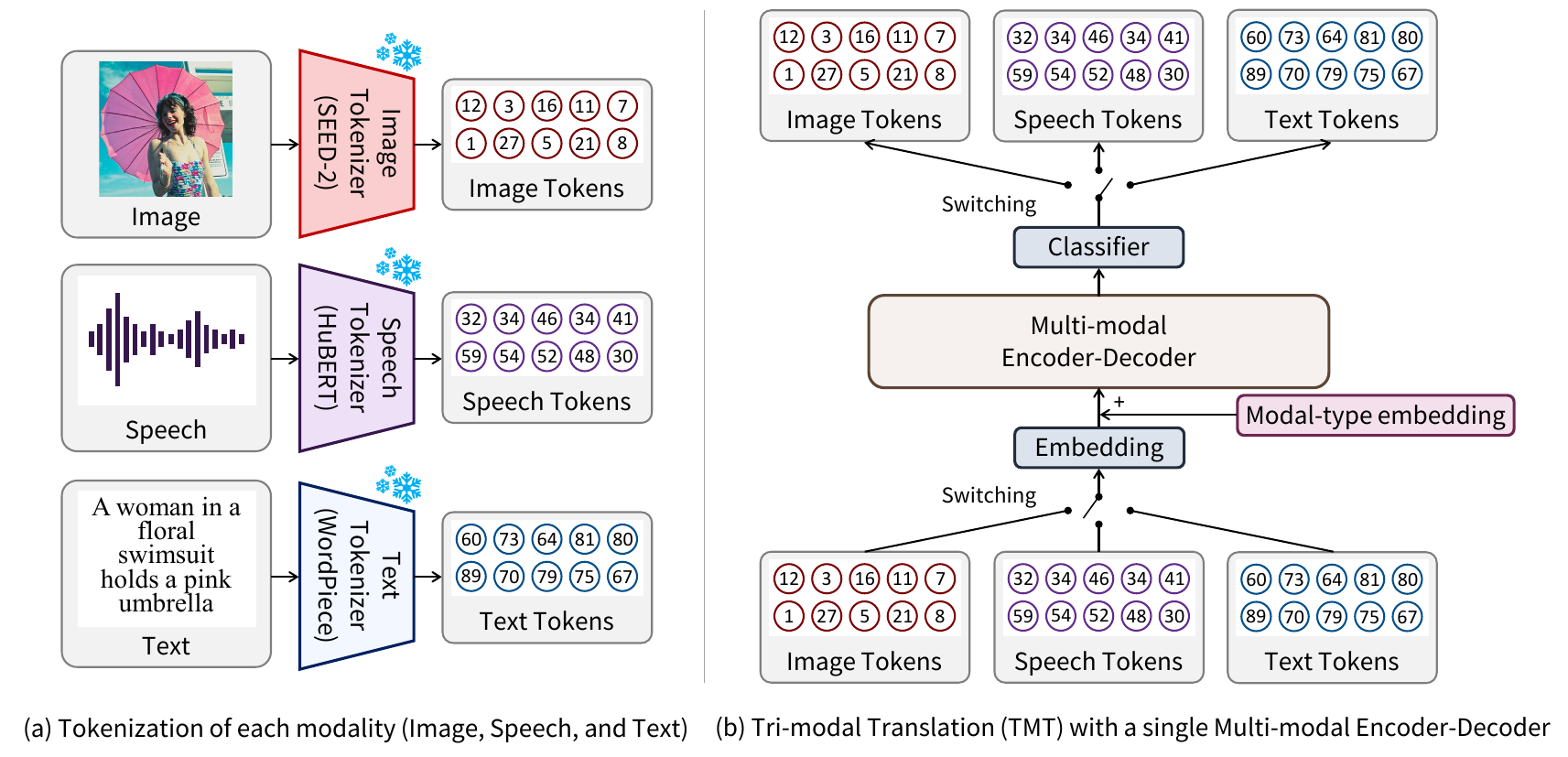}}
    \vspace{-0.3cm}
	\caption{Illustration of the overall proposed Tri-modal Translation (TMT) model. 
 (a) Image, speech, and text modalities are tokenized into discrete tokens, respectively, using modality-specific pre-trained tokenizers. 
 (b) The TMT model encodes any of the three modalities and translates the encoded representation into a desired output modality. 
 All modalities are represented in discrete tokens and hence our ``{\em multi-modal}'' translation (MMT) task is treated as a neural machine translation (NMT) task between different ``{\em languages}'', with improved efficiency.
	}
	\label{fig:1}
    \vspace{-0.6cm}
\end{figure*}

\section{Method}
Fig.~\ref{fig:1} illustrates the overall scheme of the proposed TMT framework. Let $x_i$ be an image, $x_s$ be speech describing the image, and $x_t$ be the transcription of the speech. The main objective is to build a single model that freely translates between different modalities (\ie, $x_t$, $x_s$, and $x_i$). For this goal, we first tokenize image, speech, and text data into discretized tokens (Fig.~\ref{fig:1}a and $\S$\ref{ssec:individual_tokenization}). The vocabulary set of TMT is a union set of three sets of vocabularies. Then we treat discretized tokens from different modalities as different languages and train the TMT model like training an NMT model ($\S$\ref{ssec:integrated_modelling}). The proposed TMT can translate between tri-modalities, covering all six MMT tasks ($\S$\ref{ssec:tasks_covered}).

\subsection{Individual tokenization of each modality}
\label{ssec:individual_tokenization}
Discretized representations have demonstrated promising results in various research fields. In image processing, Vector Quantization (VQ)~\cite{van2017neural,esser2021taming,ge2023planting,ge2023making} has successfully expressed images using discrete tokens. In speech processing, VQ~\cite{defossez2022encodec,wang2023neural} and Self-Supervised Learning (SSL)~\cite{baevski2020wav2vec,hsu2021hubert,chen2022wavlm} model-based discrete tokens~\cite{lakhotia2021generative} are showing remarkable performances across different tasks (\eg, speech translation~\cite{lee2022textless,popuri2022enhanced,choi2023av2av} and synthesis~\cite{polyak2021speech,hayashi2020discretalk}). 

Motivated by the recent success, we tokenize image, speech, and text modalities. Specifically, as shown in Fig.~\ref{fig:1}a, the raw image is tokenized into image tokens $u_i\in\mathbb{N}^{L_i}$ by using SEED-2~\cite{ge2023making} tokenizer, where each image is compressed into 32 tokens (\ie, $L_i=32$). Speech is also tokenized into discrete values $u_s\in\mathbb{N}^{L_s}$ by clustering the features of a HuBERT~\cite{hsu2021hubert}, an SSL speech model, whose output is 50Hz. Following the practices in ~\cite{lakhotia2021generative}, we merge duplicated tokens so that the granularity of speech tokens (\ie, $L_s$) is less than 50Hz. For the text input, we employ the BERT tokenizer~\cite{schuster2012japanese,kenton2019bert} to derive text tokens $u_t\in\mathbb{N}^{L_t}$.

\textbf{Advantages of tokenizing tri-modalities.}
Tokenizing all three modalities brings two main advantages.
Firstly, the MMT problem now has a unified interface. Modality specifics are processed mostly within tokenization and detokenization process. The other key advantage is efficiency~\cite{chang2023exploration,park2023storage}. Processing raw image and audio is burdensome due to the necessity of large-scale data storage and computational resources~\cite{kim2024multilingual}. Discretizing speech and image reduces the bits required to 0.2\% and 0.035\%, which are similar levels to text data. Table \ref{tab:bits} shows the data size comparisons in bits between different data types of each modality. By employing speech tokens rather than directly using raw audio or its Mel-spectrogram, we significantly reduce the data size to less than 0.2\% compared to other formats. Consequently, 1-second speech tokens contain a similar number of bits as text comprising approximately 50 characters. Furthermore, by utilizing image tokens instead of raw images, only 0.035\% of the bits are needed to represent the original image content. This substantial reduction equates one image to the text length of 52 characters. Therefore, by representing and employing the tokenized data instead of directly utilizing their raw format, we can efficiently train our TMT like training a text-only system with reduced data storage requirements and computing resources.

\begin{table}[!t]
	\renewcommand{\arraystretch}{1.2}
	\renewcommand{\tabcolsep}{1.4mm}
\centering
\caption{Data size (bits) comparisons between raw data and tokens. 
The values are based on the 16kHz 16bits for raw speech waveforms,  224$\times$224 for raw RGB image, 80 filterbanks with 10ms shift for mel-spectrogram.
$L$ is the length (sec) of the audio. 
For text, one character is assigned 8 bits. 
$S$ refers to the length of the characters.}
\resizebox{0.9999\linewidth}{!}{
\begin{tabular}{cccc}
\Xhline{3\arrayrulewidth}
\textbf{Modality} & \textbf{Data Format} & \textbf{Data Size (bits)} & \textbf{Data Size (\%)}\\ \hline
\multirow{3}{*}{\textbf{Speech}} & Raw Audio &  16000 $\times$ L $\times$ 16 & 100\% \\
& Mel-spectrogram & 100 $\times$ 80 $\times$ L $\times$ 32 & 100\% \\ 
& Speech Token & ($<$50) $\times$ L $\times$ 8 & $<$0.2\% \\ \hline
\multirow{2}{*}{\textbf{Image}} & Raw Image & 224 $\times$ 224 $\times$ 3 $\times$ 8 & 100\% \\
& Image Token & 32 $\times$ 13  & 0.035\%\\ 
\hline
\textbf{Text} & Raw Text & S $\times$ 8 & - \\
\Xhline{3\arrayrulewidth}
\end{tabular}}
\label{tab:bits}
\end{table}

\subsection{Integrated modeling of tri-modalities}
\label{ssec:integrated_modelling}
As shown in Fig.~\ref{fig:1}b, we leverage a multi-modal encoder-decoder architecture~\cite{vaswani2017attention,lewis2020bart,liu2020multilingual}, widely used in the NMT literature as the backbone architecture. Therefore, all parameters are shared between all six MMT tasks. A popular auto-regressive training scheme is employed. To inform both the encoder and the decoder about input and output modalities, we add a modal-type embedding to the embedded features after the token embedding layer, similar to language embeddings~\cite{conneau2019cross}. Then, the losses for each MMT are summed and backpropagated. Formally put, the objective function of TMT training can be expressed as:
\begin{equation}
\setlength{\abovedisplayskip}{4pt}
\setlength{\belowdisplayskip}{4pt}
\label{eq:1}
\mathcal{L} = - \sum_{\substack{m\in\mathcal{M}\\ m\neq k}} \ \sum_{k\in\mathcal{M}} \ \sum^{L_m}_{l=1} \ \log p(u_m^l|u_m^{<l},u_k),
\end{equation}
where $k$ and $m$ refer to a pair of non-overlapping modalities, $\mathcal{M}\,$=$\,\{i,s,t\}$ is the set of tri-modalities, and $u^{<l}$ represents the predictions before $l$-th frame.

\begin{figure*}[h!]
\centering
\centerline{\includegraphics[width=0.902\textwidth]{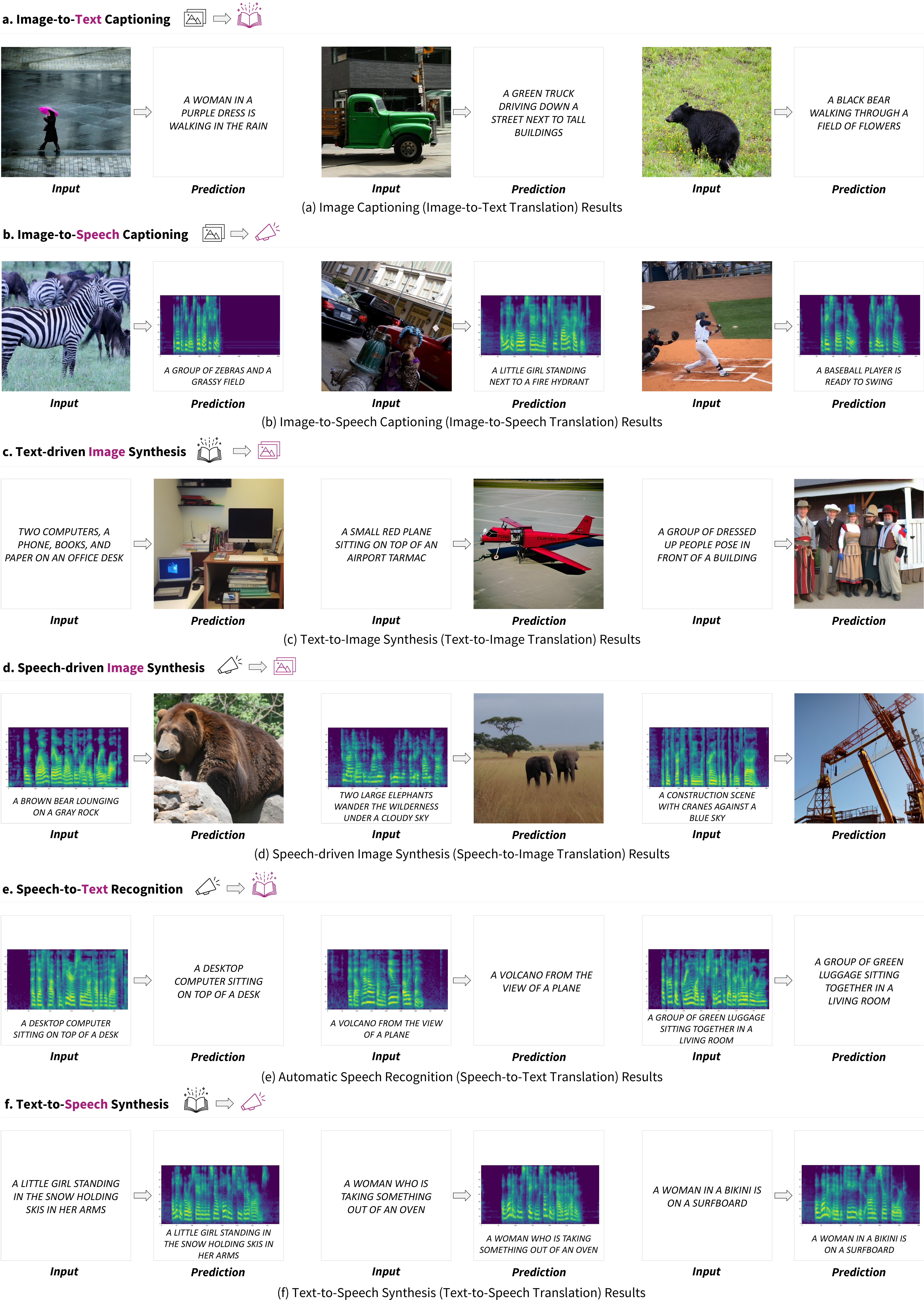}}
\caption{
  Sample illustrations of the six tasks that the proposed TMT is capable of. (a) Image captioning, (b) Image-to-speech captioning, (c) Text-to-image synthesis, (d) Speech-to-image synthesis, (e) Automatics speech recognition, and (f) Text-to-speech synthesis. Note that we also show transcriptions, obtained using a pre-trained ASR system, of speech to ease understanding. 
}
\label{fig:2}
\end{figure*}

\subsection{MMT tasks covered by TMT}
\label{ssec:tasks_covered}
TMT incorporates all six MMT tasks ($_3\mathbf{P}_2$) spanning different combinations of input and output modalities.

\textbf{Image Captioning} 
\cite{xu2015show} is a task of describing input image content using natural language. A comprehensive understanding of the input image and the accuracy and completeness of the output text are important.

\textbf{Image-to-Speech Captioning} \cite{hsu2021text} is a task that directly synthesizes a spoken description of the input image. 
This task is used for the image-to-speech translation evaluation of TMT. 
It is regarded as a challenging task due to inherent complex nature of speech where speech contains not only linguistic information but also diverse factors of acoustic noise, different tones, delivery fast, etc.

\textbf{Text-driven Image Synthesis} \cite{ramesh2021dalle,ramesh2022hierarchical,saharia2022photorealistic,chang2023muse} aims to synthesize an image that matches a given text description. Both the naturalness of the generated image and its correctness to the input description are important.

\textbf{Speech-driven Image Synthesis} generates an image from a spoken description~\cite{li2020direct,wang2020s2igan}. The task is equal to solving an automatic speech recognition (ASR) and text-driven image synthesis at once without the intermediate text representation. Similar to the relationship between image captioning and image-to-speech captioning, further challenges exist, mainly due to the absence of text modality. This task remains crucial for those languages that do not have writing systems~\cite{lee2022textless}, where text-based systems cannot be applied. Note that this paper is the first to synthesize high-resolution images of 768 $\times$ 768 from speech.

\textbf{ASR} is a well-established task that generates text transcriptions from an input speech~\cite{amodei2016deep,prabhavalkar2023end}. We use this task for speech-to-text translation of TMT.

\textbf{Text-to-Speech Synthesis (TTS)} generates speech corresponding to the given text transcription~\cite{wang2017tacotron,tan2023neural}.

\subsection{Back Translation (BT)}
\label{ssec:bt}
We investigate whether applying BT in the MMT problem based on our interpretation can help improve the performance even if utilizing uni-modal databases. To this end, TMT model is first pre-trained with approximately 10M paired tri-modal data (detailed in $\S$\ref{sec:dataset}) to produce an intermediate model. Then, with the intermediate model, we apply BT on a uni-modal image dataset, ImageNet~\cite{deng2009imagenet}, to produce pseudo source modalities of speech and text. This results in 2.6M text-image and speech-image data. We also apply BT on a speech-text bi-modal dataset, CommonVoice~\cite{ardila2020common}, and produce additional 2M image-speech and image-text data. By merging the pseudo data on top of the original tri-modal data pairs, we further train the intermediate TMT model and derive the final TMT model. Please note that, we employ the intermediate TMT model trained on tri-modal paired data to perform BT, instead of using other image captioning or ASR models.

\begin{table*}[t]
	\renewcommand{\arraystretch}{1.4}
	\renewcommand{\tabcolsep}{2.4mm}
\caption{
Comparisons between single MMT systems and TMT on the image and speech-driven captioning tasks.
}
\centering
\resizebox{0.999\linewidth}{!}{
\begin{tabular}{ll ccccc ccccc}
\Xhline{3\arrayrulewidth}
\multirow{2}{*}{\textbf{Task}} & \multirow{2}{*}{\textbf{Methods}} & \multicolumn{5}{c}{\textbf{COCO}} & \multicolumn{5}{c}{\textbf{Flickr8k}} 
\\ \cmidrule(l{2pt}r{2pt}){3-7} \cmidrule(l{2pt}r{2pt}){8-12}
& & \textbf{BLEU-4} & \textbf{METEOR} & \textbf{ROUGE} & \textbf{CIDEr} & \textbf{SPICE}
& \textbf{BLEU-4} & \textbf{METEOR} & \textbf{ROUGE} & \textbf{CIDEr} & \textbf{SPICE} \\ \cmidrule(l{2pt}r{2pt}){1-2} \cmidrule(l{2pt}r{2pt}){3-7} \cmidrule(l{2pt}r{2pt}){8-12}

\multirow{2}{*}{\textbf{Image $\rightarrow$ Text}} 
& Single MMT & 28.9 & 27.3 & 53.6 & 101.9 & 20.5 & 21.8 & 24.3 & 49.7 & 67.7 & 19.1 \\
& \textbf{TMT} & \textbf{31.4} & \textbf{28.1} & \textbf{54.9} & \textbf{108.7} & \textbf{21.3} & \textbf{26.0} & \textbf{25.8} & \textbf{52.5} & \textbf{79.7} & \textbf{20.5} \\
\cmidrule(l{2pt}r{2pt}){1-2} \cmidrule(l{2pt}r{2pt}){3-7} \cmidrule(l{2pt}r{2pt}){8-12}
\multirow{2}{*}{\textbf{Image $\rightarrow$ Speech}} 
& Single MMT & 22.5 & 21.5 & 47.1 & 69.1 & 15.0 & 16.9 & 20.0 & 45.6 & 44.7 & 13.5 \\
& \textbf{TMT} & \textbf{24.7} & \textbf{23.2} & \textbf{48.9} & \textbf{78.7} & \textbf{16.7} & \textbf{21.3} & \textbf{22.2} & \textbf{48.5} & \textbf{55.2} & \textbf{15.4} \\
\Xhline{3\arrayrulewidth}
\end{tabular}}
\label{tab:single_multi_cap}
\vspace{-0.2cm}
\end{table*}
\begin{table}[t]
	\renewcommand{\arraystretch}{1.4}
	\renewcommand{\tabcolsep}{2.5mm}
\centering
\caption{
Comparisons between single MMT systems and TMT on four MMT tasks.
}
\resizebox{0.9999\linewidth}{!}{
\begin{tabular}{ll cc cc}
\Xhline{3\arrayrulewidth}
\multirow{2}{*}{\textbf{Task}} & \multirow{2}{*}{\textbf{Methods}} & \multicolumn{2}{c}{\textbf{COCO}} & \multicolumn{2}{c}{\textbf{Flickr8k}} 
\\ \cmidrule(l{2pt}r{2pt}){3-4} \cmidrule(l{2pt}r{2pt}){5-6}
& & \textbf{CLIP} & \textbf{WER}
& \textbf{CLIP} & \textbf{WER} \\ \cmidrule(l{2pt}r{2pt}){1-2} \cmidrule(l{2pt}r{2pt}){3-4} \cmidrule(l{2pt}r{2pt}){5-6}

\multirow{2}{*}{\textbf{Text $\rightarrow$ Image}} 
& Single MMT & \textbf{68.8} & - & \textbf{69.3} & - \\
& \textbf{TMT} & 68.2 & - & 68.1 & - \\
\cmidrule(l{2pt}r{2pt}){1-2} \cmidrule(l{2pt}r{2pt}){3-4} \cmidrule(l{2pt}r{2pt}){5-6}
\multirow{2}{*}{\textbf{Speech $\rightarrow$ Image}} 
& Single MMT & 57.5 & - & 58.5 & - \\
& \textbf{TMT} & \textbf{67.3} & - & \textbf{67.0} & - \\
\cmidrule(l{2pt}r{2pt}){1-2} \cmidrule(l{2pt}r{2pt}){3-4} \cmidrule(l{2pt}r{2pt}){5-6}

\multirow{2}{*}{\textbf{Speech $\rightarrow$ Text}} 
& Single MMT & - & 6.9 & - & 6.8 \\
& \textbf{TMT} & - & \textbf{6.5} & - & \textbf{6.1} \\
\cmidrule(l{2pt}r{2pt}){1-2} \cmidrule(l{2pt}r{2pt}){3-4} \cmidrule(l{2pt}r{2pt}){5-6}
\multirow{2}{*}{\textbf{Text $\rightarrow$ Speech}} 
& Single MMT & - & 16.0 & - & 9.7 \\
& \textbf{TMT} & - & \textbf{11.8} & - & \textbf{9.5} \\
\Xhline{3\arrayrulewidth}
\end{tabular}}
\label{tab:single_multi_other}
\end{table}

\section{Experiments}
\subsection{Datasets}
\label{sec:dataset}
The training data comprises Conceptual Captions 3M (CC3M), Conceptual Captions 12M (CC12M)~\cite{sharma2018conceptual,changpinyo2021conceptual}, COCO~\cite{lin2014coco}, SpokenCOCO~\cite{hsu2021text}, Flickr8k~\cite{hodosh2013flickr}, and Flickr8kAudio~\cite{harwath2015flickraudio}. 

\textbf{CC3M} is an image-text paired dataset that consists of about 3.3M paired instances. Different from the curated style of other image caption annotations, the dataset is collected from the web through an automatic pipeline. Therefore, the image-text pairs represent a wider variety of styles.

\textbf{CC12M} is an image-text paired dataset with about 12M instances, which is adequate for pretraining vision-language systems. The dataset is similarly collected as CC3M with relaxed constraints during its automatic pipeline to further increase the data size.

\textbf{COCO} is a popular dataset for various computer vision tasks. The data provides images with diverse annotations including object segmentations, object bounding boxes, object classes, key points, and text descriptions. We employed the images and corresponding text descriptions to train image-to-text and text-to-image tasks of TMT.

\textbf{SpokenCOCO} is the speech version of COCO dataset. The dataset is collected by recording human speakers reading the text annotations of COCO. By merging SpokenCOCO with COCO, we can build speech-image-text tri-modal paired datasets which makes it possible to train all 6 tasks of TMT.

\textbf{Flickr8k} is a famous image captioning dataset contains 8K images with five different captions for each image. The images are collected from six different Flickr groups to contain diverse scenes and situations.

\textbf{Flickr8kAudio} is the speech version of the Flickr8k dataset, similar to the relationship between COCO and SpokenCOCO. The dataset is constructed by recording the speech from 183 speakers. By merging Flickr8kAudio with Flickr8k dataset, we can build speech-image-text tri-modal paired datasets.

For CC3M and CC12M, VITS~\cite{kim2021vits}, a TTS model trained on VCTK \cite{yamagishi2019vctk}, is employed to synthesize speech from random speakers to compile audio-text-image pairs. The evaluation is performed on the test split of COCO and Flickr8k after finetuning. The popular Karpathy splits \cite{karpathy2015deep} are employed for COCO and Flickr8k.

\subsection{Metrics}
We leverage diverse metrics to evaluate six MMT tasks spanning three output modalities.

\textbf{Captioning tasks.}
For the image captioning and image-to-speech captioning tasks, we employ BLEU~\cite{papineni2002bleu}, METEOR~\cite{denkowski2014meteor}, ROUGE~\cite{lin2004rouge}, CIDEr~\cite{vedantam2015cider}, and SPICE~\cite{anderson2016spice}. BLEU, METEOR, ROUGE, and CIDEr measure n-gram overlaps between predictions and references. SPICE compares semantic propositional content. To measure these metrics for the speech outputs (\ie, image-to-speech captioning), we employ a pre-trained ASR model~\cite{baevski2020wav2vec} to transcribe the speech into text following \cite{kim2024towards}. Then, the metric is measured on text-level. Across all metrics, higher value indicates better performance.

\textbf{Image synthesis tasks.}
For the text- and speech-driven image synthesis tasks, we adopt the CLIP score~\cite{radford2021clip} (\ie, the cosine similarity) to assess how similar the generated image is associated with the input text or speech, following~\cite{ge2023planting,koh2023generating}.

\textbf{ASR and TTS.}
We employ the widely used Word Error Rate (WER) for ASR and TTS. For TTS, we transcribe the generated speech by using a pre-trained ASR model~\cite{baevski2020wav2vec}, following~\cite{lakhotia2021generative,kim2023many}. Additionally, we employ Mean Opinion Score (MOS) and neural MOS to assess the quality of generated results.

\begin{table*}[t]
	\renewcommand{\arraystretch}{1.4}
	\renewcommand{\tabcolsep}{1.5mm}
\centering
\caption{Image captioning performance (Image-to-Text Translation) comparisons on COCO and Flickr8k. 
We also report the performance of raw image-based models for reference purposes. 
``Num. MMT.'' indicates the number of MMT tasks that the model is capable of.}
\resizebox{\linewidth}{!}{
\begin{tabular}{ccc ccccc ccccc}
\Xhline{3\arrayrulewidth}
\multirow{2}{*}{\makecell{\textbf{Input}\\ \textbf{Representation}}} & \multirow{2}{*}{\textbf{Methods}} & \multirow{2}{*}{\makecell{\textbf{Num.}\\ \textbf{MMT.}}} & \multicolumn{5}{c}{\textbf{COCO}} & \multicolumn{5}{c}{\textbf{Flickr8k}} 
\\ \cmidrule(l{2pt}r{2pt}){4-8} \cmidrule(l{2pt}r{2pt}){9-13}
& & & \textbf{BLEU-4} & \textbf{METEOR} & \textbf{ROUGE} & \textbf{CIDEr} & \textbf{SPICE} 
& \textbf{BLEU-4} & \textbf{METEOR} & \textbf{ROUGE} & \textbf{CIDEr} & \textbf{SPICE} \\ \cmidrule(l{2pt}r{2pt}){1-3} \cmidrule(l{2pt}r{2pt}){4-8} \cmidrule(l{2pt}r{2pt}){9-13}

\multirow{4}{*}{\makecell{Raw Image}} & SAT \cite{xu2015show} & 1 & 24.3 & 23.9 & - & - & - & 21.3 & 20.3 & - & - & - \\
& LaDic \cite{wang2024ladic} & 1 & 
38.2 & 29.5 & 58.7 & 123.2 & 22.4 & - & - & - & - & - \\
& Kim~\etal~\cite{kim2024towards,wang2022git} & 1 & 
38.7 & 29.5 & 59.1 & 131.2 & 23.3 & 30.8 & 26.9 & 55.8 & 93.8 & 20.0 \\ 
& BLIP-2 \cite{li2023blip2} & 1 & 
38.5 & 32.5 & 60.1 & 135.8 & 26.4 & 30.4 & 29.4 & 56.4 & 99.6 & 22.5 \\ 
 \cmidrule(l{2pt}r{2pt}){1-3} \cmidrule(l{2pt}r{2pt}){4-8} \cmidrule(l{2pt}r{2pt}){9-13}

\multirow{3}{*}{\makecell{\textbf{Discrete}\\\textbf{Image Token}}} 
& Kim~\etal~\cite{kim2024towards} & 1 & 29.9 & 25.2 & 52.8 & 97.4 & 18.6 & 23.4 & 22.0 & 48.9 & 63.3 & 15.4 \\
& Ge~\etal~\cite{ge2023planting,ge2023making} & 2 & \textbf{34.6} & 28.4 & 56.4 & \textbf{131.0} & \textbf{22.9} & 24.4 & 25.3 & 51.7 & 75.4 & 18.1 \\ \cdashline{2-13}
& \textbf{TMT} & 6 & 34.3 & \textbf{29.2} & \textbf{56.9} & 117.7 & 22.3 & \textbf{29.7} & \textbf{26.5} & \textbf{54.8} & \textbf{90.6} & \textbf{20.9} \\
\Xhline{3\arrayrulewidth}
 \end{tabular}}
\label{table:img2txt}
\vspace{-0.1cm}
\end{table*}
\begin{table*}[t]
	\renewcommand{\arraystretch}{1.4}
	\renewcommand{\tabcolsep}{1.5mm}
\centering
\caption{Image-to-speech captioning performance (Image-to-Speech Translation) comparisons on COCO and Flickr8k. 
We also report the performance of raw image-based models for reference purposes. ``Num. MMT.'' indicates the number of MMT tasks that the model is capable of.
}
\resizebox{\linewidth}{!}{
\begin{tabular}{ccc ccccc ccccc}
\Xhline{3\arrayrulewidth}
\multirow{2}{*}{\makecell{\textbf{Input}\\ \textbf{Representation}}} & \multirow{2}{*}{\textbf{Methods}} & \multirow{2}{*}{\makecell{\textbf{Num.}\\ \textbf{MMT.}}} & \multicolumn{5}{c}{\textbf{COCO}} & \multicolumn{5}{c}{\textbf{Flickr8k}} 
\\ \cmidrule(l{2pt}r{2pt}){4-8} \cmidrule(l{2pt}r{2pt}){9-13}
& & & \textbf{BLEU-4} & \textbf{METEOR} & \textbf{ROUGE} & \textbf{CIDEr} & \textbf{SPICE} 
& \textbf{BLEU-4} & \textbf{METEOR} & \textbf{ROUGE} & \textbf{CIDEr} & \textbf{SPICE} \\ \cmidrule(l{2pt}r{2pt}){1-3} \cmidrule(l{2pt}r{2pt}){4-8} \cmidrule(l{2pt}r{2pt}){9-13}

\multirow{4}{*}{\makecell{Raw Image}} 
& Wang~\etal~\cite{wang2021synthesizing} & 1 & - & - & - & - & - & 3.5 & 11.3 & 23.2 & 8.0 & - \\
& Hsu~\etal~\cite{hsu2021text} & 1 & 23.3 & 21.2 & 47.8 & 73.2 & 14.9& 12.5 & 14.5 & 39.1 & 24.5 & 9.5  \\
& Effendi~\etal~\cite{effendi2021end} & 1 & - & - & - & - & - & 14.8 & 17.4 & 32.9 & 45.8 & - \\
& Kim~\etal~\cite{kim2024towards} & 1 & 25.9 & 23.8 & 50.4 & 81.1 & 17.5 & 20.6 & 22.0 & 48.4 & 53.6 & 15.8 \\
\cmidrule(l{2pt}r{2pt}){1-3} \cmidrule(l{2pt}r{2pt}){4-8} \cmidrule(l{2pt}r{2pt}){9-13}

\multirow{2}{*}{\makecell{\textbf{Discrete}\\\textbf{Image Token}}} 
& Kim~\etal~\cite{kim2024towards} & 1 
& 20.1 & 21.4 & 46.4 & 64.0 & 15.0 & 16.7 & 19.6 & 44.2 & 41.2 & 13.1 \\ \cdashline{2-13}
& \textbf{TMT} & 6 & \textbf{25.8} & \textbf{23.3} & \textbf{49.5} & \textbf{81.3} & \textbf{16.8} & \textbf{21.1} & \textbf{21.9} & \textbf{48.6} & \textbf{54.9} & \textbf{15.3} \\
\Xhline{3\arrayrulewidth}
\end{tabular}}
\label{tab:img2sp}
\vspace{-0.1cm}
\end{table*}

\subsection{Implementation Details}
\label{ssec:expcfg}
The input image is resized into 224 $\times$ 224, while the output image size is 768 $\times$ 768. The input audio is resampled to 16kHz and tokenized using a pre-trained HuBERT base model~\cite{hsu2021hubert}. The backbone model that models the multi-modal information is an encoder-decoder Transformer~\cite{vaswani2017attention}. Both the encoder and decoder are initialized from a pre-trained BERT-Base configuration~\cite{kenton2019bert}, unless otherwise specified. Therefore, each encoder and decoder contains 12 layers, where each layer has an embedding dimension of $768$, a feed-forward dimension of $3072$, and $12$ attention heads. If BERT-Large configuration is applied, it comprises 24 layers with an embedding dimension of $1024$, a feed-forward dimension of $4096$, and $16$ heads.
 
Each token is embedded through an embedding layer, where the total vocabulary size is $38914$ consisting of $8192$, $200$, and $30522$ vocabularies from image, speech, and text respectively. 
We add a modality type embedding that signifies which modality the input belongs to in the same manner as adding positional embeddings. 
We use speech samples up to a length of 384 tokens after deduplicating speech tokens. During the pre-training phase before applying BT, we use a batch size of 80 for each MMT task, a peak learning rate of $1e^{-4}$ with warm-up, Adam optimizer~\cite{kingma2014adam}, and 350k steps. 
Then, with the intermediate pre-trained TMT model, BT is applied on ImageNet and CommonVoice as described in $\S$\ref{ssec:bt}. Then, we further train the model using the additional data for 250k steps. Finally, we finetune the model for 50k steps on COCO and Flickr8k with a peak learning rate of $1e^{-5}$.
For the detokenization, we employ the decoder of SEED-2~\cite{ge2023making} for image and a token-based HiFi-GAN vocoder~\cite{kong2020hifi,lee2022direct} trained on LJSpeech~\cite{ljspeech17}, for speech. To handle the removed duration information in the speech tokens during the deduplication, the vocoder is augmented with the duration predictor as in \cite{ren2021fastspeech,lee2022direct}. Therefore, the speech token is repeated according to the predicted duration before generating the raw waveform. For decoding, we use beam search with a beam width of 5. Models are trained on 4 A40 (48GB) GPUs.

\subsection{Results}
Firstly, we show qualitative results for each task of TMT in Fig.~\ref{fig:2}. In the next sections, we compare single MMT models with the proposed TMT model. Then, we evaluate TMT against existing literature. We also conduct ablation studies on tokenizers, translation module, and the impact of using different modalities. Lastly, we include a human subjective evaluation on Image-to-Speech synthesis and TTS tasks.

\subsubsection{Effectiveness of Unifying Tasks}
\label{ssec:single}
We start with the question: {\em ``Can training a single model, capable of multiple MMT tasks, have synergy in performance on top of efficiency compared to training multiple single MMT models?''} We train six single MMT models and one TMT model using the train sets of COCO and Flickr8k. For fairness, all models are trained for 300k iterations.

Table~\ref{tab:single_multi_cap} compares the captioning performances, including image-to-text and image-to-speech translations. The results show that by unifying tasks, we can consistently obtain performance gain. In our analysis, this stems from the unified multi-modal encoder-decoder. The encoder-decoder can leverage complementary multi-modal information in learning language modeling, as both speech and text modalities contain linguistic information. Table~\ref{tab:single_multi_other} shows the comparison results for the remaining four tasks. 
Similar to captioning, the unified TMT model outperforms its single-task counterparts consistently, except for the text-to-image translation. However, the performance differences between single MMT and TMT in text-to-image translation are marginal, while the performance gain obtained by unifying tasks is substantial for speech-to-image translation. We conclude that unifying tasks is beneficial in terms of both practicality (\ie, by building a single model) and performance. More detailed analysis on different task combinations can be found in $\S$\ref{ssec:diffmodal}.

\subsubsection{Comparison with the literature}
We compare TMT's performance on each MMT task with the previous works.

\textbf{Image Captioning Tasks.}
Table~\ref{table:img2txt} shows the results of image-to-text translation of TMT and other competing systems. We also report the performances of raw image-based systems for reference purposes. We observe that our TMT accomplishes comparable performances among state-of-the-art discrete image-based models. Notably, the proposed TMT is competitive with the LLM-based method~\cite{ge2023planting} with a much smaller number of parameters. Note that while the modality translation module  (excluding the tokenizer and detokenizer) of \cite{ge2023planting} is built on OPT 2.7B model~\cite{zhang2022opt}, the proposed TMT employs a 270M multi-modal model. This shows the effectiveness of the proposed TMT compared to the method of finetuning LLM. Moreover, TMT can perform six MMT tasks while the previous methods can only perform two MMT tasks at maximum.

Table~\ref{tab:img2sp} shows the direct image-to-speech captioning performances. TMT outperforms the recently proposed state-of-the-art~\cite{kim2024towards} on both COCO and Flickr8k databases and even achieves comparable performances with raw-image based method. Please note that the vocoder used in \cite{kim2024towards} and TMT is the same. Since the decoder of the proposed TMT is trained using multi-tasks with text simultaneously, it better models linguistic information compared to the transfer learning from the text of \cite{kim2024towards}.

\textbf{Image Synthesis Tasks.}
The upper part of Table~\ref{tab:imsyn_asr_tts} displays the performances of text- and speech-driven image synthesis tasks. Following the literature, we evaluate the model on Flickr30k instead of using Flickr8k for the text-to-image synthesis tasks. The proposed TMT achieves the best CLIP similarity on both databases in text-to-image translation, indicating that TMT generates images that best describe the given text sentences. Please note that TMT still outperforms the competitors with more diverse MMT task abilities. Also, the speech-to-image translation performance of TMT on COCO is 68.88, close to that of text-to-image performance, while the cascaded system of \cite{rombach2022ldm} and Wav2Vec2.0 (W2V2) \cite{baevski2020wav2vec} achieves 62.87. 
We confirm that even if using speech directly as input, TMT correctly generates images described in the speech. We underline that this work is the first to explore high-resolution image synthesis at a resolution of 768$\times$768 from speech. 

\textbf{ASR and TTS Tasks.}
The lower part of Table~\ref{tab:imsyn_asr_tts} shows the performances of ASR and TTS. For evaluation purposes, we compare the ASR performances with pre-trained WavLM~\cite{chen2022wavlm}, HuBERT~\cite{hsu2021hubert}, and, W2V2~\cite{baevski2020wav2vec} models. The results show that TMT can correctly recognize the input speech into text by achieving lower WER compared to other pre-trained methods. For TTS, as the proposed model is a discrete token-based speech synthesis model, we compare the performances with UTUT~\cite{kim2023many} and the resynthesized speech (ReSyn) from the ground-truth speech tokens following~\cite{lakhotia2021generative,kim2023many}. Please note that the vocoder used for ReSyn and TMT is the exact same model; the only difference between the two is how the input speech tokens for the vocoder are obtained. ReSyn employs the ground truth speech tokens, while TMT employs speech tokens that predicted by itself from the input text. Since the vocoder is exactly same, we can access how the speech tokens produced by TMT are in a good quality by comparing its performance with that of the resynthesis. Results demonstrate that even though the proposed TMT is trained using ground truth speech tokens (which inherently contain errors), it effectively learns the relationships between text and speech by suppressing undesired noises present in the ground truth speech tokens, and generates correct speech from input text.

\begin{table}[t]
	\renewcommand{\arraystretch}{1.4}
	\renewcommand{\tabcolsep}{3mm}
\centering
\caption{
Text- and Speech-driven image synthesis, ASR, and TTS performance comparisons on COCO and Flickr8k. 
For TTS, we report the WER by feeding the synthesized audio into a pre-trained ASR model.
}
\resizebox{\linewidth}{!}{
\begin{tabular}{cccc}
\Xhline{3\arrayrulewidth}
\multirow{2}{*}{\textbf{Methods}} & \multirow{2}{*}{\makecell{\textbf{Num.}\\ \textbf{MMT.}}} & \multicolumn{2}{c}{\textbf{Image Synthesis (CLIP)}} \\ \cmidrule(l{2pt}r{2pt}){3-4}
& & \textbf{COCO} & \textbf{Flickr30k} \\ \cmidrule(l{2pt}r{2pt}){1-2} \cmidrule(l{2pt}r{2pt}){3-3} \cmidrule(l{2pt}r{2pt}){4-4}
\multicolumn{4}{l}{$\bullet$ \textbf{\textit{Text-driven Image Synthesis}}} \\
LDM \cite{rombach2022ldm} & 1 & 68.43 & 65.40 \\
Muse \cite{chang2023muse,patil2024amused} & 1 & 62.45 & 58.40 \\
Koh~\etal~\cite {koh2023generating} & 2 & 67.45 & 65.16 \\
UniDiffuser \cite{bao2023one} & 2 & 67.26 & 63.00 \\
Ge~\etal~\cite{ge2023planting} & 2 & 68.23 & 65.23 \\ 
 \hdashline
\textbf{TMT} & 6 & \textbf{69.57} & \textbf{66.32} \\ 
\hline
\multicolumn{4}{l}{$\bullet$ \textbf{\textit{Speech-driven Image Synthesis}}} \\
Cascaded System \cite{baevski2020wav2vec,rombach2022ldm} & 1 + 1 & 62.87 & - \\ \hdashline
\textbf{TMT} & 6 & \textbf{68.88} & - \\ 
\Xhline{3\arrayrulewidth}

\multicolumn{4}{c}{} \\ [-3ex]
\Xhline{3\arrayrulewidth}
\multirow{2}{*}{\textbf{Methods}} & \multirow{2}{*}{\makecell{\textbf{Num.}\\ \textbf{MMT.}}} & \multicolumn{2}{c}{\textbf{Speech Processing (WER)}} \\ \cmidrule(l{2pt}r{2pt}){3-4}
& & \textbf{COCO} & \textbf{Flickr8k} \\ \cmidrule(l{2pt}r{2pt}){1-2} \cmidrule(l{2pt}r{2pt}){3-3} \cmidrule(l{2pt}r{2pt}){4-4}
\multicolumn{4}{l}{$\bullet$ \textbf{\textit{Automatic Speech Recognition (ASR)}}} \\
WavLM \cite{chen2022wavlm} & 1 & 16.69 & 11.68  \\ 
HuBERT \cite{hsu2021hubert} & 1 & 10.14 & 7.51 \\ 
W2V2 \cite{baevski2020wav2vec} & 1 & 9.70 & 7.26 \\ \hdashline
\textbf{TMT} & 6 & \textbf{5.15} & \textbf{4.60} \\ 
\hline
\multicolumn{4}{l}{$\bullet$ \textbf{\textit{Text-to-Speech Synthesis (TTS)}}} \\
ReSyn \cite{lakhotia2021generative} & - & 29.32 & 22.21 \\ 
UTUT \cite{kim2023many} & 1 & 12.06 & 12.75 \\ 
\hdashline
\textbf{TMT} & 6 & \textbf{9.54} & \textbf{8.46} \\ 
\Xhline{3\arrayrulewidth}
\end{tabular}}
\label{tab:imsyn_asr_tts}
\vspace{-0.2cm}
\end{table}

\subsubsection{Ablation on Speech and Image Tokens}
To confirm the effectiveness of the different multi-modal tokens, we perform ablation studies by differing configurations for speech and image tokens. First, we differentiate the number of k-means clustering of the speech tokenizer to obtain different token sizes. Then, we train single MMT models of image-to-speech captioning and ASR using 200, 500, and 1000 token sizes. Table \ref{table:unitabl} shows the image-to-speech captioning and ASR performances. The results show that there is a trade-off in performances of ASR (WER) and image-to-speech captioning (BLEU, METEOR, ROUGE, CIDEr, SPICE).
We confirm a consistent tendency where image-to-speech captioning performs better with a smaller number of speech tokens.
In contrast, for ASR, employing a large number of tokens is beneficial. 
The trade-off in number of tokens can be understood as follows: since speech tokens are produced through quantization, the number of tokens is directly related to the level of information compression. Specifically, employing a higher number of speech tokens allows for the utilization of finer-grained speech features~\cite{lakhotia2021generative,chang2023exploration}. However, the benefits of these finer-grained features depend on the task at hand. One possible intuition is that with a sufficient amount of speech tokens, the model can focus on generating speech with correct linguistic content, while an excessive number of similar-sounding tokens may cause distraction in content modeling.
On the other hand, in ASR, utilizing a large number of speech tokens enables to capture subtle pronunciation differences. Since the performance gaps in ASR are smaller than those of image-to-speech captioning, we employ 200 speech tokens for all experiments. 

Moreover, we evaluate the impact of using different SSL speech models as tokenizers on image-to-speech captioning and ASR tasks. To this end, we train the models with different tokens: one is tokenized using WavLM \cite{chen2022wavlm} following \cite{chang2023exploration}, and the other is tokenized using HuBERT following \cite{lakhotia2021generative}. Similar to the previous experiment, we train single MMT models of image-to-speech captioning and ASR using each tokenizer. Table~\ref{table:speechfeature} shows the comparison between the two tokenizers. We note that using HuBERT as the speech tokenizer achieves better performances than using WavLM.

\begin{table}[t]
	\renewcommand{\arraystretch}{1.4}
	\renewcommand{\tabcolsep}{1.3mm}
\centering
\caption{Ablation study to confirm the effects of vocabulary size of speech tokens on COCO.}
\resizebox{0.9999\linewidth}{!}{
\begin{tabular}{l cccccc}
\Xhline{3\arrayrulewidth}
\textbf{\# Token} & \textbf{BLEU-4} & \textbf{METEOR} & \textbf{ROUGE} & \textbf{CIDEr} & \textbf{SPICE} & \textbf{WER} \\ \cmidrule(l{2pt}r{2pt}){1-1} \cmidrule(l{2pt}r{2pt}){2-7}
\textbf{200 Tokens} & \textbf{22.5} & \textbf{21.5} & \textbf{47.1} & \textbf{69.1} & \textbf{15.0} & 6.9 \\
\textbf{500 Tokens} & 17.7 & 19.0 & 44.3 & 52.7 & 12.3 & \textbf{6.3} \\
\textbf{1000 Tokens} & 15.1 & 17.7 & 41.7 & 48.3 & 11.9 & \textbf{6.3} \\
\Xhline{3\arrayrulewidth}
\end{tabular}}
\label{table:unitabl}
\vspace{-0.1cm}
\end{table}

Second, we investigate the different types of visual tokens among VQ-GAN-based \cite{esser2021taming,yu2021vector} and SEED-based \cite{ge2023planting,ge2023making}. To check the effect of visual tokens only, we employ the same architecture and training configuration of \cite{kim2024towards} and only change the visual tokens. The performances of image captioning and image-to-speech captioning on COCO are shown in Table \ref{table:visualunitabl}. In both tasks, SEED-based token achieves better performances. As SEED tokenizer is trained along with image-text association learning, its token achieves better performance in image-language association with fewer tokens (\ie, 784~vs.~32). Therefore, we employ SEED-based tokenizer for our experiments.

\begin{table}[t]
	\renewcommand{\arraystretch}{1.4}
	\renewcommand{\tabcolsep}{1.3mm}
\centering
\caption{Performance comparisons using different SSL speech models as tokenizers on COCO.}
\resizebox{0.9999\linewidth}{!}{
\begin{tabular}{l cccccc}
\Xhline{3\arrayrulewidth}
\textbf{Token Type} & \textbf{BLEU-4} & \textbf{METEOR} & \textbf{ROUGE} & \textbf{CIDEr} & \textbf{SPICE} & \textbf{WER} \\ \cmidrule(l{2pt}r{2pt}){1-1} \cmidrule(l{2pt}r{2pt}){2-7}
WavLM \cite{chen2022wavlm} & 8.6 & 13.5 & 34.9 & 21.3 & 6.9 & 15.9 \\
HuBERT \cite{hsu2021hubert} & \textbf{22.5} & \textbf{21.5} & \textbf{47.1} & \textbf{69.1} & \textbf{15.0} & \textbf{6.9} \\
\Xhline{3\arrayrulewidth}
\end{tabular}}
\label{table:speechfeature}
\vspace{-0.1cm}
\end{table}

\begin{table}[t]
	\renewcommand{\arraystretch}{1.4}
	\renewcommand{\tabcolsep}{2.0mm}
\centering
\caption{Captioning tasks' performance comparisons using different visual token types on COCO.}
\resizebox{0.9999\linewidth}{!}{
\begin{tabular}{l ccccc}
\Xhline{3\arrayrulewidth}
\multirow{2}{*}{\makecell{\textbf{Visual}\\ \textbf{Token Type}}} & \multicolumn{5}{c}{\textbf{Image Captioning}} \\ \cmidrule(l{2pt}r{2pt}){2-6}
& \textbf{BLEU-4} & \textbf{METEOR} & \textbf{ROUGE} & \textbf{CIDEr} & \textbf{SPICE} \\ \cmidrule(l{2pt}r{2pt}){1-1} \cmidrule(l{2pt}r{2pt}){2-6}
VQ-GAN-based & 29.9 & 25.2 & 52.8 & 97.4 & 18.6 \\
SEED-based & \textbf{33.0} & \textbf{27.1} & \textbf{55.5} & \textbf{117.2} & \textbf{20.7} \\ \Xhline{3\arrayrulewidth}
\multicolumn{6}{c}{} \\ [-3.0ex]
\Xhline{3\arrayrulewidth}
\multirow{2}{*}{\makecell{\textbf{Visual}\\ \textbf{Token Type}}} & \multicolumn{5}{c}{\textbf{Image-to-Speech Captioning}} \\ \cmidrule(l{2pt}r{2pt}){2-6}
& \textbf{BLEU-4} & \textbf{METEOR} & \textbf{ROUGE} & \textbf{CIDEr} & \textbf{SPICE} \\ \cmidrule(l{2pt}r{2pt}){1-1} \cmidrule(l{2pt}r{2pt}){2-6}
VQ-GAN-based & 20.1 & 21.4 & 46.4 & 64.0 & 15.0 \\
SEED-based & \textbf{20.6} & \textbf{21.6} & \textbf{47.4} & \textbf{65.1} & \textbf{15.7} \\
\Xhline{3\arrayrulewidth}
\end{tabular}}
\label{table:visualunitabl}
\vspace{-0.1cm}
\end{table}

\subsubsection{Ablation Study on Multi-modal Translation Module}
We perform ablation study by using different models for the multi-modal encoder-decoder module. Firstly, we investigate different initializations of translation module using BART \cite{lewis2020bart}, mBART \cite{liu2020multilingual}, and BERT \cite{kenton2019bert}. Specifically, the models are initialized using one of the pre-trained models and trained on 6 tasks for 300k steps on COCO and Flickr8k dataset. The comparison results on image captioning and image-to-speech captioning are shown in Table~\ref{table:bertbartabl}. Both BART and BERT initialized models achieve high performances on two captioning tasks, while multilingual pre-trained model, mBART achieves inferior performances on both tasks. This might be contributed to the fact that COCO and Flickr8k are English only databases. Considering the performance gap between BERT and BART initialized  models in image captioning task is much bigger than that of image-to-speech captioning, we use BERT for our encoder-decoder initialization.

Furthermore, we analyze the effects of using different model sizes for the translation module. To this end, we compare four different types of translation modules of 1) BERT-base for both encoder-decoder, 2) BERT-base for encoder and BERT-large for decoder, 3) BERT-large for encoder and BERT-base for decoder, and 4) BERT-large for both encoder-decoder. BERT-base has 12 layers with 110M parameters, while BERT-large has 24 layers with 340M parameters. All models are trained on full training set including back translation. The comparison results can be found in Table~\ref{table:translator}. Interestingly, different models achieve the best performance on different tasks. For example, model with BERT-large for both encoder-decoder (\ie, Large-Large) achieves the best result on image-to-speech captioning task while showing inferior image synthesis performances. The model with BERT-large for encoder and BERT-base for decoder (\ie, Large-Base) achieves the best image synthesis performances while showing inferior image-to-speech captioning task. Scaling up the model size shows a tendency of performance improvement when we compare the performance with the BERT-base model for both encoder-decoder (\ie, Base-Base). Given that the number of parameters of Base-Base model is much smaller than others (\ie, Base-Base: 270M parameters, Large-Large: 800M parameters) and still showing comparable performances on all 6 tasks, we report its performance for other experiments.

\begin{table}[t]
	\renewcommand{\arraystretch}{1.4}
	\renewcommand{\tabcolsep}{2.0mm}
\centering
\caption{Captioning tasks' performance comparisons using different encoder-decoder initializations on COCO.}
\resizebox{0.9999\linewidth}{!}{
\begin{tabular}{l ccccc}
\Xhline{3\arrayrulewidth}
\multirow{2}{*}{\makecell{\textbf{Translation}\\\textbf{Module}}} & \multicolumn{5}{c}{\textbf{Image Captioning}} \\ \cmidrule(l{2pt}r{2pt}){2-6}
& \textbf{BLEU-4} & \textbf{METEOR} & \textbf{ROUGE} & \textbf{CIDEr} & \textbf{SPICE} \\ \cmidrule(l{2pt}r{2pt}){1-1} \cmidrule(l{2pt}r{2pt}){2-6}
BART \cite{lewis2020bart} & 28.9 & 26.6 & 53.2 & 103.2 & 20.4 \\
mBART \cite{liu2020multilingual} & 28.4 & 26.3 & 53.1 & 100.9 & 19.9 \\ 
BERT \cite{kenton2019bert} & \textbf{31.4} & \textbf{28.1} & \textbf{54.9} & \textbf{108.7} & \textbf{21.3} \\
\Xhline{3\arrayrulewidth}
\multicolumn{6}{c}{} \\ [-3.0ex]
\Xhline{3\arrayrulewidth}
\multirow{2}{*}{\makecell{\textbf{Translation}\\\textbf{Module}}} & \multicolumn{5}{c}{\textbf{Image-to-Speech Captioning}} \\ \cmidrule(l{2pt}r{2pt}){2-6}
& \textbf{BLEU-4} & \textbf{METEOR} & \textbf{ROUGE} & \textbf{CIDEr} & \textbf{SPICE} \\ \cmidrule(l{2pt}r{2pt}){1-1} \cmidrule(l{2pt}r{2pt}){2-6}
BART \cite{lewis2020bart} & \textbf{24.9} & \textbf{23.4} & \textbf{49.9} & \textbf{80.8} & \textbf{17.2} \\
mBART \cite{liu2020multilingual} & 22.6 & 22.6 & 48.1 & 71.8 & 15.9 \\ 
BERT \cite{kenton2019bert} & 24.7 & 23.2 & 48.9 & 78.7 & 16.7 \\
\Xhline{3\arrayrulewidth}
\end{tabular}}
\label{table:bertbartabl}
\vspace{-0.1cm}
\end{table}
\begin{table*}[t]
	\renewcommand{\arraystretch}{1.3}
	\renewcommand{\tabcolsep}{2.2mm}
\centering
\caption{Ablation study using different network configurations for multi-modal encoder-decoder on COCO and Flickr8k dataset.
`Base' and `Large' represents BERT-Base and BERT-Large configurations, respectively. ROUGE is reported for captioning tasks, WER for ASR and TTS, and the CLIP score for image synthesis tasks.}
\resizebox{0.85\linewidth}{!}{
\begin{tabular}{ccccccc}
\Xhline{3\arrayrulewidth}

\multirow{2}{*}{\makecell{\textbf{Models}\\\textbf{(Encoder-Decoder)}}} & \multicolumn{6}{c}{\textbf{COCO}}\\ \cline{2-7}
&  \textbf{Image Captioning$\uparrow$} & {\textbf{Image-to-Speech$\uparrow$}} & \textbf{ASR$\downarrow$} & \textbf{Text-to-Image$\uparrow$} & {\textbf{Speech-to-Image$\uparrow$}} & \textbf{TTS$\downarrow$}\\
\hline
Base-Base & 56.9 & 49.5 & 5.15 & 69.57 & 68.88 & \textbf{9.54} \\
Base-Large & \textbf{57.3} & 49.0 & \textbf{4.88} & 69.46 & 68.82 & 10.55 \\
Large-Base & 57.0 & 47.7 & 4.96 & \textbf{70.49} & \textbf{69.74} & 10.21 \\
Large-Large & 57.1 & \textbf{50.1} & 5.01 & 69.14 & 68.33 & 10.14 \\
\Xhline{3\arrayrulewidth}
\multicolumn{6}{c}{} \\ [-3.0ex]
\Xhline{3\arrayrulewidth}
\multirow{2}{*}{\makecell{\textbf{Models}\\\textbf{(Encoder-Decoder)}}} & \multicolumn{6}{c}{\textbf{Flickr8k}}\\\cline{2-7}
&  \textbf{Image Captioning$\uparrow$} & {\textbf{Image-to-Speech$\uparrow$}} & \textbf{ASR$\downarrow$} & \textbf{Text-to-Image$\uparrow$} & {\textbf{Speech-to-Image$\uparrow$}} & \textbf{TTS$\downarrow$}\\
\hline
Base-Base & 54.8 & 48.6 & 4.60 & 70.24 & 68.98 & \textbf{8.46} \\
Base-Large & \textbf{55.2} & 48.9 & \textbf{4.26} & 69.90 & 68.78 & 8.91 \\
Large-Base & 55.1 & 48.3 & 4.32 & \textbf{71.36} & \textbf{70.01} & 8.89 \\
Large-Large & 54.9 & \textbf{49.8} & 4.42 & 69.73 & 67.94 & 8.90\\
\Xhline{3\arrayrulewidth}
\end{tabular}}
\label{table:translator}
\vspace{-0.1cm}
\end{table*}

\begin{table}[t]
	\renewcommand{\arraystretch}{1.3}
	\renewcommand{\tabcolsep}{2.5mm}
\centering
\caption{Effectiveness of Back Translation (BT) on COCO. ROUGE is reported for captioning tasks. 
    }
\resizebox{0.9999\linewidth}{!}{
\begin{tabular}{cccc}
\Xhline{3\arrayrulewidth}

\textbf{Methods} & \textbf{Image Captioning$\uparrow$} & {\textbf{Image-to-Speech$\uparrow$}} & \textbf{ASR$\downarrow$}\\
\hline
TMT w/o BT & 56.7 & 48.9& 5.24\\
\textbf{TMT} & \textbf{56.9} & \textbf{49.5}& \textbf{5.15}\\
\Xhline{3\arrayrulewidth}

\textbf{Methods} & \textbf{Text-to-Image$\uparrow$} & {\textbf{Speech-to-Image$\uparrow$}} & \textbf{TTS$\downarrow$}\\
\hline
TMT w/o BT & 69.48 & 68.67 & 9.78\\
\textbf{TMT} & {\bf 69.57}& \textbf{68.88} &\textbf{9.54}\\
\Xhline{3\arrayrulewidth}
\end{tabular}}
\label{tab:bt}
\vspace{-0.1cm}
\end{table}
\begin{table}[t]
	\renewcommand{\arraystretch}{1.3}
	\renewcommand{\tabcolsep}{2.0mm}
\centering
\caption{Human evaluation score (MOS) comparisons with 95\% confidence interval, and Neural MOS scores in image-to-speech captioning and TTS on COCO.}
\resizebox{0.999\linewidth}{!}{
\begin{tabular}{ccccc}
\Xhline{3\arrayrulewidth}
\multirow{2}{*}{\textbf{Methods}} & \multicolumn{2}{c}{\textbf{MOS}} & \multicolumn{2}{c}{\textbf{Neural MOS}} \\ \cmidrule(l{2pt}r{2pt}){2-3} \cmidrule(l{2pt}r{2pt}){4-5}
& \textbf{Natural} & \textbf{Descriptive} & \textbf{MOSNet} $\uparrow$ & \textbf{SpeechLM} $\downarrow$ \\ \cmidrule(l{2pt}r{2pt}){1-3} \cmidrule(l{2pt}r{2pt}){4-5}
\multicolumn{5}{l}{$\bullet$ \textbf{\textit{Image-to-Speech Captioning}}} \\
Hsu~\etal~\cite{hsu2021text} & 2.70$_{\pm 0.38}$ & 2.96$_{\pm 0.22}$ & 2.51 & 4.23 \\
Kim~\etal~\cite{kim2024towards} & 3.71$_{\pm 0.25}$ & 3.41$_{\pm 0.23}$ & 4.30 & 4.16 \\ \hdashline
\textbf{TMT} & \textbf{3.82}$_{\pm 0.26}$ & \textbf{3.60}$_{\pm 0.21}$ & \textbf{4.31} & \textbf{4.07} \\ \hline
\multicolumn{5}{l}{$\bullet$ \textbf{\textit{Text-to-Speech Synthesis (TTS)}}} \\
VITS \cite{kim2021vits} & \textbf{3.85}$_{\pm 0.25}$ & - & \textbf{4.64} & \textbf{4.02} \\ \hdashline
\textbf{TMT} & 3.77$_{\pm 0.36}$ & - & 4.31 & 4.11 \\
\Xhline{3\arrayrulewidth}
\end{tabular}}
\label{tab:mos}      
\vspace{-0.1cm}
\end{table}

\subsubsection{Effectiveness of Back Translation}
Table~\ref{tab:bt} shows the result of training TMT with and without BT. By employing additional 4M back translated data, we observe consistent improvement across tasks of TMT. This shows the potential for further improvement, since the amount of additional data BT can produce is limitless.

\begin{table}[t]
	\renewcommand{\arraystretch}{1.4}
	\renewcommand{\tabcolsep}{0.5mm}
\centering
\caption{
Ablation study on using different modalities with different MMT tasks. $S,I,T$ represent Speech, Image, and Text modalities, respectively, and arrow represents the translation directions. Best is bolded and second-best is underlined.
}
\resizebox{0.9999\linewidth}{!}{
\begin{tabular}{ll ccc ccc}
\Xhline{3\arrayrulewidth}
\multirow{2}{*}{\textbf{Task}} & \multirow{2}{*}{\makecell{\textbf{Training}\\ \textbf{Methods}}} & \multicolumn{3}{c}{\textbf{COCO}} & \multicolumn{3}{c}{\textbf{Flickr8k}} 
\\ \cmidrule(l{2pt}r{2pt}){3-5} \cmidrule(l{2pt}r{2pt}){6-8}
& & \textbf{ROUGE$\uparrow$} & \textbf{CLIP$\uparrow$} & \textbf{WER$\downarrow$}
& \textbf{ROUGE$\uparrow$} & \textbf{CLIP$\uparrow$} & \textbf{WER$\downarrow$} \\ \cmidrule(l{2pt}r{2pt}){1-2} \cmidrule(l{2pt}r{2pt}){3-5} \cmidrule(l{2pt}r{2pt}){6-8}

\multirow{5}{*}{\textbf{I $\rightarrow$ T\,\,}} 
& I $\rightarrow$ T                     & 53.6 & - & - & 49.7 & - & - \\
& I $\rightarrow$ T + I $\rightarrow$ S & \underline{54.1} & - & - & 50.3 & - & - \\
& I $\rightarrow$ T + S $\rightarrow$ T & \underline{54.1} & - & - & \underline{51.1} & - & - \\
& I $\rightarrow$ T + S $\rightarrow$ I & 52.0 & - & - & 46.1 & - & - \\
& \textbf{TMT (all 6 tasks)} & \textbf{54.9} & - & - & \textbf{52.5} & - & - \\
\cmidrule(l{2pt}r{2pt}){1-2} \cmidrule(l{2pt}r{2pt}){3-5} \cmidrule(l{2pt}r{2pt}){6-8}

\multirow{5}{*}{\textbf{I $\rightarrow$ S\,\,}} 
& I $\rightarrow$ S                     & 47.1 & - & - & 45.6 & - & - \\
& I $\rightarrow$ S + I $\rightarrow$ T & \textbf{49.3} & - & - & \underline{47.7} & - & - \\
& I $\rightarrow$ S + T $\rightarrow$ S & 42.1 & - & - & 40.2 & - & - \\
& I $\rightarrow$ S + T $\rightarrow$ I & 46.6 & - & - & 45.3 & - & - \\
& \textbf{TMT (all 6 tasks)} & \underline{48.9} & - & - & \textbf{48.5} & - & - \\
\cmidrule(l{2pt}r{2pt}){1-2} \cmidrule(l{2pt}r{2pt}){3-5} \cmidrule(l{2pt}r{2pt}){6-8}

\multirow{5}{*}{\textbf{T $\rightarrow$ I\,\,}} 
& T $\rightarrow$ I                     & - & \underline{68.8} & - & - & \textbf{69.3} & - \\
& T $\rightarrow$ I + T $\rightarrow$ S & - & 68.6 & - & - & \textbf{69.3} & - \\
& T $\rightarrow$ I + S $\rightarrow$ I & - & 68.3 & - & - & \underline{68.2} & - \\
& T $\rightarrow$ I + S $\rightarrow$ T & - & \textbf{69.0} & - & - & \textbf{69.3} & - \\
& \textbf{TMT (all 6 tasks)} & - & 68.2 & - & - & 68.1 & - \\
\cmidrule(l{2pt}r{2pt}){1-2} \cmidrule(l{2pt}r{2pt}){3-5} \cmidrule(l{2pt}r{2pt}){6-8}

\multirow{5}{*}{\textbf{S $\rightarrow$ I\,\,}} 
& S $\rightarrow$ I                     & - & 57.5 & - & - & 58.5 & - \\
& S $\rightarrow$ I + S $\rightarrow$ T & - & 67.1 & - & - & \underline{66.6} & - \\
& S $\rightarrow$ I + T $\rightarrow$ I & - & \underline{67.2} & - & - & 66.5 & - \\
& S $\rightarrow$ I + T $\rightarrow$ S & - & 49.0 & - & - & 49.6 & - \\
& \textbf{TMT (all 6 tasks)} & - & \textbf{67.3} & - & - & \textbf{67.0} & - \\
\cmidrule(l{2pt}r{2pt}){1-2} \cmidrule(l{2pt}r{2pt}){3-5} \cmidrule(l{2pt}r{2pt}){6-8}

\multirow{5}{*}{\textbf{S $\rightarrow$ T\,\,}} 
& S $\rightarrow$ T                     & - & - & 6.9 & - & - & 6.8 \\
& S $\rightarrow$ T + S $\rightarrow$ I & - & - & \textbf{6.2} & - & - & \textbf{5.7} \\
& S $\rightarrow$ T + I $\rightarrow$ T & - & - & 6.6 & - & - & 6.3 \\
& S $\rightarrow$ T + I $\rightarrow$ S & - & - & \textbf{6.2} & - & - & \textbf{5.7} \\
& \textbf{TMT (all 6 tasks)} & - & - & \underline{6.5} & - & - & \underline{6.1} \\
\cmidrule(l{2pt}r{2pt}){1-2} \cmidrule(l{2pt}r{2pt}){3-5} \cmidrule(l{2pt}r{2pt}){6-8}

\multirow{5}{*}{\textbf{T $\rightarrow$ S\,\,}} 
& T $\rightarrow$ S                     & - & - & 16.0 & - & - & 9.7 \\
& T $\rightarrow$ S + T $\rightarrow$ I & - & - & 14.9 & - & - & \textbf{9.2} \\
& T $\rightarrow$ S + I $\rightarrow$ S & - & - & \textbf{11.1} & - & - & 11.3 \\
& T $\rightarrow$ S + I $\rightarrow$ T & - & - & 12.5 & - & - & \textbf{9.2} \\
& \textbf{TMT (all 6 tasks)} & - & - & \underline{11.8} & - & - & \underline{9.5} \\

\Xhline{3\arrayrulewidth}
\end{tabular}}
\label{tab:different_modality}
\end{table}

\subsubsection{Mean Opinion Score Comparisons}
\label{sec:mos}
In order to assess the quality of generated speech in both image-to-speech captioning and text-to-speech tasks, we conduct human subjective study and neural MOS test using MOSNet \cite{lo2018mosnet} and SpeechLMScore \cite{maiti2023speechlmscore}. Specifically, we enlisted 21 participants who are volunteers mostly in academia to evaluate the naturalness of the generated speech for both tasks, and for the image-to-speech captioning task, we also assessed descriptiveness, gauging the extent to which the speech accurately describes the input images. Participants were asked to evaluate each sample using a score between 1 and 5, with increments of 0.5. Following the previous works \cite{hsu2021text,kim2024towards}, 20 samples are used for the image-to-speech captioning task. For TTS, 20 samples and 100 samples are employed for human subjective study and neural MOS, respectively. The results are shown in Table \ref{tab:mos}. TMT achieves the best score for both human study and neural MOS scores in image-to-speech captioning. Moreover, we achieve comparable performances with VITS on TTS task. Please note that the TTS performance of TMT can’t help relying on the performance of the token-based vocoder (\ie, HiFi-GAN \cite{kong2020hifi}). As VITS showed better performances than HiFi-GAN-based TTS model in general~\cite{kim2021vits}, this tendency could be reflected in our experiment, where TMT shows slightly worse MOS scores compared to VITS. This means that there is the possibility of the improvement in terms of naturalness of the generated speech of TMT by improving the speech token-based vocoder. Yet, through MOS evaluation, we can confirm the effectiveness of TMT in generating intelligible and natural speech. 

\subsubsection{Analysis on the impact of using different modalities}
\label{ssec:diffmodal}
We have shown that the TMT trained on 6 MMT tasks with 3 modalities performs better than its single MMT counterparts in $\S$\ref{ssec:single}. In this part, we further analyze how using different modalities affect to the performance for each task. To this end, we start by building a single MMT model trained using two modalities out of three. Then, we add the third modality but with different task compositions as follows: 1) tasks with the same input modality, 2) tasks with the same output modality, and 3) tasks with no overlap in modalities both input and output. For example as an image-to-text translation (I$\rightarrow$T) task, we compare the models trained with the tasks of I$\rightarrow$T + I$\rightarrow$S (\ie, same input modality tasks), I$\rightarrow$T + S$\rightarrow$T (\ie, same output modality tasks), and I$\rightarrow$T + S$\rightarrow$I (\ie, no overlap in both input and output), by adding extra speech modality (\ie, S). Comparison results for all 6 MMT tasks are shown in Table~\ref{tab:different_modality}. All models are trained on COCO and Flickr8k for 300k steps. We report ROUGE for captioning tasks, CLIP score for image synthesis tasks, and WER for ASR and TTS tasks. The experimental results demonstrate that different tasks get benefit from different combinations of training methods. For example, on image captioning and speech-to-image synthesis tasks (\ie, I$\rightarrow$T and S$\rightarrow$I), employing an extra modality consistently improves performance except in the case where the additional task has no overlap both input and output modalities with the target task. However, in text-to-image translation and speech-to-text translation (\ie, T$\rightarrow$I and S$\rightarrow$T), employing additional task with no overlap modality in input and output achieves the best performance. Therefore, to achieve the best performance for each specific task by adding different modalities, we need to carefully select which tasks to combine. Another fact we can confirm is that speech and text modalities are complementary, so that when we utilize both of them together as input or output modalities, we can achieve the better performances in general. Moreover, employing an extra modality with a suitable extra task brings improvement in performance for the target task. Finally, by training the model with all possible tasks (\ie, 6 tasks), we can achieve a high-performance system for all 6 tasks in general. This has the advantage where it does not require extra effort to select extra tasks carefully, and a single trained model can perform all 6 tasks, showing the practicality of the proposed TMT.

\section{Conclusion}
\label{sec:conclu}
We introduced TMT, a simple yet efficient and effective tri-modal translation model between speech, image, and text modalities. We discretized different modalities and treat them as text. Our experiments demonstrated that the three modalities could be successfully translated using a unified multi-modal encoder-decoder architecture, incorporating six MMT tasks in a single model. TMT outperformed single MMT model counterparts. Remarkably, we showed the computational efficiency of TMT by reducing the training data size (in bits) to about 0.2\% for speech and 0.04\% for image data. 

 
\bibliographystyle{IEEEtran}
\bibliography{main}


\end{document}